\documentclass[11pt,a4paper]{article}
\usepackage[hyperref]{acl2020}
\usepackage{times}
\usepackage{latexsym}

\usepackage{times}
\usepackage{latexsym}
\usepackage{multirow}
\usepackage{url}
\usepackage{graphicx}
\usepackage{bm}
\usepackage{amsmath}
\usepackage{amssymb}
\usepackage{amstext}
\usepackage{mathrsfs}
\usepackage{array}
\usepackage{booktabs}

\usepackage{color}

\usepackage{microtype}

\aclfinalcopy % Uncomment this line for the final submission
\title{Align, Mask and Select: A Simple Method for Incorporating \\ Commonsense Knowledge into Language Representation Models}

\author{Zhi-Xiu Ye,\textsuperscript{\rm 1}\thanks{~~Work was done during an internship at DAMO Academy, Alibaba Group.} Qian Chen,\textsuperscript{\rm 2} Wen Wang,\textsuperscript{\rm 2}  Zhen-Hua Ling\textsuperscript{\rm 1} \\
\textsuperscript{\rm 1}National Engineering Laboratory for Speech and Language Information Processing, \\
University of Science and Technology of China \\
\textsuperscript{\rm 2}Speech Lab, DAMO Academy, Alibaba Group \\
zxye@mail.ustc.edu.cn, tanqing.cq@alibaba-inc.com, \\
w.wang@alibaba-inc.com, zhling@ustc.edu.cn
}
\date{}

\begin{document}
\maketitle
\begin{abstract}
The state-of-the-art pre-trained language representation models, such as Bidirectional Encoder Representations from Transformers (BERT), rarely incorporate commonsense knowledge or other knowledge explicitly. We propose a pre-training approach for incorporating commonsense knowledge into language representation models. We construct a commonsense-related multi-choice question answering dataset for pre-training a neural language representation model. The dataset is created automatically by our proposed ``align, mask, and select" (AMS) method. We also investigate different pre-training tasks.
Experimental results demonstrate that pre-training models using the proposed approach followed by fine-tuning achieve significant improvements over previous state-of-the-art models on two commonsense-related benchmarks, including CommonsenseQA and Winograd Schema Challenge. We also observe that fine-tuned models after the proposed pre-training approach maintain comparable performance on other NLP tasks, such as sentence classification and natural language inference tasks, compared to the original BERT models. These results verify that the proposed approach, while significantly improving commonsense-related NLP tasks, does not degrade the general language representation capabilities.

% Neural language representation models can well capture rich language information from unlabeled text, and can be fine-tuned to benefit many NLP applications.
% However, the existing pre-trained language representation models rarely consider explicitly incorporating commonsense knowledge or other knowledge.
% In this paper, we propose a pre-training approach for incorporating commonsense knowledge into language representation models. 
% We construct a commonsense-related multi-choice question answering dataset for pre-training a neural language representation model.
% The dataset is created automatically by our proposed ``align, mask, and select" (AMS) method. 
% % We also investigate different pre-training tasks.
% Experimental results demonstrate that pre-training models using the proposed approach followed by fine-tuning achieve significant improvements over previous state-of-the-art models on various commonsense-related tasks, such as CommonsenseQA and Winograd Schema Challenge. 
% We also find that fine-tuned models maintain comparable performance on other NLP tasks, such as sentence classification and natural language inference tasks, compared to the original language representation models.
%These results verify that the proposed approach, while significantly improving commonsense-related NLP tasks, does not degrade the language representation capabilities.

\end{abstract}

\section{Introduction}

Recently, significant progress has been made in language representation models \cite{pennington-etal-2014-glove,peters-etal-2017-semi,howard-ruder-2018-universal,radford2018improving,devlin-etal-2019-bert,zhang2019ernie}. These models can be categorized into feature-based approaches and fine-tuning approaches. In particular, a pre-training/fine-tuning technique, Bidirectional Encoder Representations from Transformers (BERT) \cite{devlin-etal-2019-bert}, was proposed and has quickly created state-of-the-art models for a wide variety of NLP tasks such as question answering (QA), text classification, and natural language inference (NLI) \cite{rajpurkar-etal-2016-squad,wang-etal-2018-glue}.

% Neural language representation models, such as \citep{pennington-etal-2014-glove,peters-etal-2017-semi,howard-ruder-2018-universal,radford2018improving,devlin-etal-2019-bert,zhang2019ernie}, have been shown effective for capturing semantics from text and consistently improve the performance of various downstream natural language processing (NLP) tasks.
% Bidirectional Encoder Representations from Transformers (BERT) \citep{devlin-etal-2019-bert}, as one of the most recently developed models, has produced the state-of-the-art (SOTA) results by fine-tuning on various NLP tasks, including natural language inference (NLI) \citep{bowman-etal-2015-large}, named entity recognition (NER) 
%\citep{tjong-kim-sang-de-meulder-2003-introduction}, question answering  \citep{rajpurkar-etal-2016-squad,zellers-etal-2018-swag}, text classification \citep{wang-etal-2018-glue}, and has achieved human-level performances on several datasets \citep{rajpurkar-etal-2016-squad,zellers-etal-2018-swag}.

\begin{table}[t]
	\small
	\begin{center}
		\begin{tabular}{p{7.3cm}}
			\toprule
			A) Some examples from CSQA dataset \\
			\midrule
			What can eating lunch cause that is painful? \\
			headache, gain weight, farts, bad breath, \textbf{heartburn} \\
			\midrule
			What is the main purpose of having a bath? \\
			\textbf{cleanness}, water, exfoliation, hygiene, wetness \\
			\midrule
			Where could you find a shark before it was caught? \\
            business, marine museum, pool hall, \textbf{tomales bay}, desert \\
			\bottomrule
			\toprule
			B) Some triples from ConceptNet \\
			\midrule
			(eating dinner, Causes, heartburn) \\
			(eating dinner, MotivatedByGoal, not get headache) \\
			(lunch, Synonym, dinner) \\
			\midrule
			(have bath, HasSubevent, cleaning) \\
			\midrule
			(shark, AtLocation, tomales bay) \\
			\bottomrule
		\end{tabular}
	\end{center}
	\caption{Some examples from the CSQA dataset shown in Part A and some related triples from ConceptNet shown in Part B. The correct answers in Part A are in boldface.}
	\label{tab:example}
\end{table}

However, commonsense reasoning remains a challenging task for modern machine learning methods. 
For example, recently \citet{talmor-etal-2019-commonsenseqa} proposed a commonsense-related task, \textit{CommonsenseQA}(CSQA), and showed that the BERT model accuracy remains dozens of points lower than the human accuracy on questions about commonsense knowledge.
Some examples from CSQA are shown in Part A of Table \ref{tab:example}. As can be seen from the examples, although it is easy for humans to answer the questions based on their world knowledge, it is a great challenge for machines when there is limited training data. We hypothesize that exploiting knowledge graphs (KGs) representing commonsense in QA modeling may help the model choose correct answers. For example, as shown in Part B of Table~\ref{tab:example}, some triples from ConceptNet \citep{speer2017conceptnet} are related to the questions above. Exploiting these triples in QA modeling may help the models make the correct decision.

In this paper, we propose a pre-training approach that can leverage commmonsense KGs, such as ConceptNet \citep{speer2017conceptnet}, to improve the commonsense reasoning capability of language representation models, such as BERT, without sacrificing the language representation capabilities of the models. 
That is, we also aim to maintain comparable performances on other NLP tasks with the original BERT models. 
It is challenging to incorporate the commonsense knowledge into language representation models since the commonsense knowledge is usually represented in a structured format, such as (concept$_1$, relation, concept$_2$) in ConceptNet, which is inconsistent with the data used for pre-training language representation models.
For example, BERT is pre-trained on the BooksCorpus and English Wikipedia that are composed of unstructured natural language sentences.
To tackle this challenge, inspired by the distant supervision approach \citep{mintz-etal-2009-distant}, we propose an ``align, mask and select" (AMS) method to automatically construct a multi-choice question-answering dataset, by aligning a commonsense KG with a large text corpus and constructing natural language sentences with labeled concepts. We then replace the masked language model (MLM) and next sentence prediction (NSP) tasks used for the original BERT pre-training stage with a multi-choice question answering task based on this dataset.

%, such as CSQA and Winograd Schema Challenge (WSC) \cite{levesque2012winograd}. 
% We observe significant improvements over previous SOTA results. We also fine-tune and evaluate the pre-trained models on other NLP tasks, such as sentence classification and NLI tasks in GLUE \cite{wang-etal-2018-glue}, and achieve comparable performance with the original BERT models.

In summary, our contributions are threefold.
\textbf{First}, we propose a pre-training approach for incorporating commonsense knowledge into language representation models for improving the commonsense reasoning capabilities of these models. This pre-training approach is agnostic to the language representation models. We propose the AMS method to automatically construct a multi-choice QA dataset and facilitate the proposed pre-training approach.
%The proposed approach is applicable to BERT and other models for the fine-tuning methods. The proposed approach is applicable to various KGs, including commonsense KGs, domain-specific KGs, etc.
\textbf{Second}, experiments demonstrate that the pre-trained models from the proposed approach with fine-tuning achieve significant improvements over previous SOTA models on two commonsense-related NLP benchmarks, CSQA and Winograd Schema Challenge (WSC), and maintain comparable performances on sentence classification and NLI tasks on the GLUE dataset, demonstrating that the proposed approach does not degrade the language representation capabilities of the models.
\textbf{Third}, extensive ablation analysis conducted on different data creation approaches and pre-training tasks helps shed light on pre-training strategies for incorporating commonsense knowledge.

\section{Related Work}
\subsection{Language Representation Model}
Language representation models have demonstrated effectiveness for improving many NLP tasks.
The early feature-based approaches \citep{mikolov2013distributed,pennington-etal-2014-glove,peters-etal-2018-deep} only use the pre-trained language representations as input features for other models.
In contrast, the fine-tuning approaches \citep{howard-ruder-2018-universal,radford2018improving,devlin-etal-2019-bert,yang2019xlnet} introduce minimal task-specific parameters trained on the downstream tasks while fine-tuning pre-trained parameters. 
Recently, there are works incorporating entity knowledge \citep{zhang2019ernie,sun2019ernie} and embedding multiple knowledge bases \citep{peters2019knowbert} into language representation models .
% Neither feature-based nor fine-tuning language representation models have incorporated the commonsense knowledge. 
In this work, we focus on incorporating commonsense knowledge in the pre-training stage.

\begin{table}[t]
	\small
	\begin{center}
		\begin{tabular}{p{7.0cm}}
			\toprule
			(1) A triple from ConceptNet \\
			\midrule
			(population, AtLocation, city) \\
			\midrule
			(2) \textbf{Align} with the English Wikipedia dataset to obtain a sentence containing ``population" and ``city" \\
			\midrule
			The largest \textbf{city} by \textbf{population} is Birmingham, which has long been the most industrialized city. \\
			\midrule
			(3) \textbf{Mask} "city" with a special token ``[QW]" \\
			\midrule
			The largest \textbf{[QW]} by \textbf{population} is Birmingham, which has long been the most industrialized city? \\
			\midrule
			4) \textbf{Select} distractors by searching (population, AtLocation, $\ast$) in ConceptNet \\
			\midrule
			(population, AtLocation, Michigan) \\
			(population, AtLocation, Petrie dish) \\
			(population, AtLocation, area with people inhabiting) \\
			(population, AtLocation, country) \\
			\midrule
			5) Generate a multi-choice question answering sample \\
			\midrule
			\textbf{question}: The largest \textbf{[QW]} by \textbf{population} is Birmingham, which has long been the most industrialized city? \\
			\textbf{candidates}: $\underline{city}$, Michigan, Petrie dish, area with people inhabiting, country \\
			\bottomrule
		\end{tabular}
	\end{center}
	\caption{
		The detailed procedure of constructing a multi-choice question answering sample with the proposed AMS method by masking concept$_2$.
		The $\ast$ in the fourth step is a wildcard character.
		The correct answer for the question is underlined.}
	\label{data_construction}
\end{table}

\subsection{Commonsense Reasoning}
\citep{petroni2019language} evaluated pretrained language models on factual and commonsense knowledge probing tasks.
\citep{zhong2018improving} ensembled commonsense knowledge based models with standard QA models and improved the commonsense reasoning ability. 
Other works directly incorporate commonsense knowledge into language representation models. 
\citet{sun2019probing} proposed to directly pre-train BERT on commonsense knowledge triples.
For any triple (concept$_1$, relation, concept$_2$), they took the concatenation of concept$_1$ and relation as the question and concept$_2$ as the correct answer.
Distractors were formed by randomly picking words or phrases in ConceptNet. 
% In this work, we also investigate directly incorporating commonsense knowledge into a unified language representation model. 
However, we hypothesize that the language representations learned in \citet{sun2019probing} may be tampered since the inputs to the model constructed this way are not natural language sentences. 
To address this issue, we propose a pre-training approach for incorporating commonsense knowledge that includes a method to construct large-scale natural language sentences.
\citet{rajani2019explain} collected the Common Sense Explanations (CoS-E) dataset and applied a Commonsense Auto-Generated Explanations (CAGE) framework to language representation models, which required a large amount of human efforts. 
In contrast, we propose the AMS method, inspired by the distant supervision approaches, to automatically construct a multi-choice QA dataset.

\subsection{Distant Supervision}
The distant supervision approach was originally proposed for generating training data for relation classification.
The approach in \citep{mintz-etal-2009-distant} assumes that if two entities/concepts participate in a relation, all sentences that mention these two entities/concepts express that relation.
It is inevitable that there exists noise in the data labeled by distant supervision \citep{riedel2010modeling}.
In this paper, instead of employing the relation labels labeled by distant supervision, we focus on the aligned entities/concepts. We propose the AMS method to construct a multi-choice QA dataset that \textbf{aligns} sentences with commonsense knowledge triples, \textbf{masks} the aligned entities/concepts in sentences and treat the masked sentences as questions,  and \textbf{selects} several entities/concepts from knowledge graphs as distractor choices.

\section{Proposed Approach}
\subsection{Commonsense Knowledge Base}
\label{sect:kb}
We use ConceptNet\footnote{\url{https://github.com/commonsense/conceptnet5/wiki}} \citep{speer2017conceptnet}, one of the most widely used commonsense knowledge bases.
ConceptNet is a semantic network that represents the large sets of words and phrases and the commonsense relationships between them (36 core relations).
It contains over 21 million edges and over 8 million nodes.
% Its English vocabulary contains approximately 1,500,000 nodes. And it contains at least 10,000 nodes for each of the 83 languages, respectively.
Each instance in ConceptNet can be represented as a triple 
%$r_i$ = 
(concept$_1$, relation, concept$_2$), indicating \textit{relation} between the two concepts \textit{concept$_1$} and \textit{concept$_2$}. 
For example, the triple (semicarbazide, IsA, chemical compound) means that ``semicarbazide is a kind of chemical compounds"; the triple (cooking dinner, Causes, cooked food) means that ``the effect of cooking dinner is cooked food", etc.

\subsection{Constructing Pre-training Dataset}
\label{sect:dataset}
We first filter the triples in ConceptNet as follows:
(1) Filter triples in which one of the concepts is not English words.
(2) Filter triples with the general relations ``RelatedTo" and ``IsA", which hold a large proportion in ConceptNet.
(3) Filter triples in which one of the concepts has more than four words or the edit distance (character-level) between the two concepts is less than four.
After filtering, we obtain 606,564 triples.
Each training sample is generated by three steps in the AMS method: \textbf{a}lign, \textbf{m}ask, and \textbf{s}elect.
Each training sample consists of a question and five candidate answers, following the form of the CSQA dataset. 

An example of constructing one training sample is shown in Table \ref{data_construction}. Firstly, we \textbf{align} each triple (concept$_1$, relation, concept$_2$) in the filtered triple set to the English Wikipedia dataset to extract the sentences containing the two concepts. This align step matches the two concepts exactly. In future work, we will explore entity linking.
Secondly, we \textbf{mask} the concept$_1$ or concept$_2$ in one sentence with a special token [QW] and treat this sentence as a question, where QW is a replacement word of the question words ``what", ``where", etc.
And the masked concept$_1$ or concept$_2$ is the correct answer for this question.
Thirdly, for generating the distractors, \citet{sun2019probing} randomly picked words or phrases in ConceptNet as the distractors. In our work, in order to generate more confusing distractors than the random selection approach, we select distractors sharing the same other unmasked concept, i.e., concept$_2$ or concept$_1$, and the same relation with the correct answer.
That is, we search ($\ast$, relation, concept$_2$) or (concept$_1$, relation, $\ast$) in ConceptNet to \textbf{select} distractors, where $\ast$ is a wildcard character that can match any word or phrase.
For each question, we reserve four distractors and one correct answer.
If there are fewer than four distractors, we discard this question.  If there are more than four distractors, we randomly select four distractors from them. After applying the AMS method, we create 16,324,846 multi-choice QA samples and denote this dataset $\mathcal{D}_{AMS}$.

\subsection{Pre-training BERT\_CS}
\label{sect:pre-training}
We explore a multi-choice QA task for pre-training the English BERT base and large models on $\mathcal{D}_{AMS}$. The resulting models are denoted BERT\_CS$_{base}$ and BERT\_CS$_{large}$, respectively. 
We then evaluate the performance of fine-tuning the BERT\_CS models on several NLP tasks (Section~\ref{sect:expts}). 

We concatenate the question with each candidate in $\mathcal{D}_{AMS}$ to construct a standard input sequence for BERT\_CS (i.e., ``[CLS] the largest [QW] by ...? [SEP] city [SEP]", where [CLS] and [SEP] are two special tokens), and the hidden representations over the [CLS] token are run through a softmax layer to predict whether the candidate is the correct answer. The objective function is defined as follows:
\begin{equation}
L = - {\rm logp}(c_i|s),
\label{objective_function}
\end{equation}
\begin{equation}
{\rm p}(c_i|s) = \frac{{\rm exp}(\mathbf{w}^{T}\mathbf{c}_{i})}{\sum_{k=1}^{N}{\rm exp}(\mathbf{w}^{T}\mathbf{c}_{k})},
\label{prob}
\end{equation}
where $c_i$ is the correct answer, $\mathbf{w}$ the parameters in the softmax layer, N the total number of candidates, and $\mathbf{c}_i$ the vector representation of the token [CLS].

To reduce the large cost of training BERT\_CS models from scratch, we initialize the BERT\_CS models (for both BERT$_{base}$ and BERT$_{large}$ models) with the parameter weights released by Google\footnote{\url{https://github.com/google-research/bert}}.
We pre-train BERT\_CS models with the batch size 160, the initial learning rate 2e-5, and the max sequence length 128 for 1 epoch. Pre-training is conducted on 16 NVIDIA V100 GPU cards with 32G memory for about 3 days for the BERT\_CS$_{large}$ model and 1 day for the BERT\_CS$_{base}$ model.

\section{Experiments}
\label{sect:expts}
Evaluations are conducted on two aspects. First, we evaluate whether the proposed approach improves the commonsense reasoning capability. Second, we investigate whether the proposed approach, while targeting improving the commonsense reasoning capability, can maintain general language representation capabilities comparable with the original BERT model, e.g., on general text classification and natural language inference (NLI) tasks. 
In the first set of experiments, we evaluate the BERT\_CS models on the benchmark CommonsenseQA (CSQA) and Winograd Schema Challenge (WSC) datasets. In the second set of experiments, we evaluate the BERT\_CS models on the General Language Understanding Evaluation (GLUE) benchmark.

\begin{table}[t!]
	\begin{center}
		\small
		\begin{tabular}{p{1.2cm}|c|c|c|c}
			\toprule
			\textbf{Dataset}             & \textbf{Train} & \textbf{Dev} & \textbf{Test} & \textbf{\#Candidates} \\
			\midrule
			CSQA       & 9741    & 1221  & 1140  & 5  \\ 
			\midrule
			WSC        & 1322    &  564  & 273   & 2  \\
			\bottomrule
		\end{tabular}
	\end{center}
	\caption{The statistics of CSQA and WSC datasets.}
	\label{dataset}
\end{table}

\begin{table}[t!]
	\begin{center}
		\small
		\begin{tabular}{p{3.8cm}|c}
			\toprule
			\textbf{Model}             & \textbf{Accuracy}  \\
			\midrule
			BERT$_{base}$     & 53.0     \\
			\midrule
			BERT$_{large}$ & 56.7 \\
			\midrule
			CoS-E \citep{rajani2019explain} & 58.2 \\
			\bottomrule
			\toprule
			BERT\_CS$_{base}$  & 56.2   \\
			\midrule
			BERT\_CS$_{large}$   & \textbf{62.2}   \\
			\bottomrule
		\end{tabular}
	\end{center}
	\caption{Accuracy (\%) of different models on the CSQA test set.}
	\label{tab:CSQA_results}
\end{table}

\begin{table*}[t!]
	\begin{center}
		\small
		\begin{tabular}{p{3.4cm}|c|c|c|c|c|c|c}
			\toprule
            \textbf{Model}                        & \textbf{WSC}   & \textbf{non-assoc.} & \textbf{assoc.}        & \textbf{unswitched} & \textbf{switched} & \textbf{consist.} & \textbf{WNLI} \\
            \midrule
            Ensemble 14 LMs                  & 63.7 & 60.6      & 83.8& 63.4 & 53.4     & 44.3     & -    \\
            \midrule
            Knowledge Hunter             & 57.1 & 58.3      & 50.0           & 58.8       &58.8     & \textbf{90.1}     & -    \\
            \midrule
            BERT$_{large}$ + MTP         & 70.3 & 70.8      & 67.6         & 73.3       &70.1     & 59.5& 70.5 \\
            \midrule
            \citet{ruan2019exploring}     & 71.1 & 69.5 & 81.1& 74.1 & 72.5 & 66.4 & - \\
            \midrule
            \citet{kocijan2019surprisingly} & 72.2 & 71.6 & 75.7 & \textbf{74.8} & 72.5 & 61.1 & 71.9 \\
            \bottomrule
            \toprule
            BERT$_{large}$ + MCQA       & 71.4 & 69.9 & 81.1 & 71.8 & 64.9 & 82.4& 78.5 \\
            \midrule
            BERT\_CS$_{large}$ + MCQA   & \textbf{75.5} & \textbf{73.7} & \textbf{86.5}&\textbf{74.8} & \textbf{73.3} & 86.3 & \textbf{83.6} \\
			\bottomrule
		\end{tabular}
	\end{center}
	\caption{Accuracy (\%) of different models on the WSC dataset together with its subsets and the WNLI test set. 
	\textbf{MTP} denotes masked token prediction, which is employed in \citet{kocijan2019surprisingly}.
	\textbf{MCQA} denotes the multi-choice question-answering format, which is employed in this paper.}
	\label{tab:WSC_results}
\end{table*}

When fine-tuning on the commonsense-related, multi-choice QA tasks, e.g., CSQA and WSC, 
we fine-tune all parameters in BERT\_CS, including the last softmax layer from the token [CLS]; whereas for the second set of experiments, we randomly initialize the classifier layer and train it from scratch. Additionally, as described in \citet{devlin-etal-2019-bert}, sometimes fine-tuning on BERT is observed to be unstable on small datasets. Hence, we run experiments with 5 different random seeds and select the best model based on the development set for all of the fine-tuning experiments in this section.

\subsection{CommonsenseQA}
\label{subsect:csqa}
The CSQA dataset consists of 12,247 questions with one correct answer and four distractors.
% This dataset consists of two splits, the question token split and the random split.
Our experiments are conducted on the more challenging random split, which is the main evaluation split \citet{talmor-etal-2019-commonsenseqa}.
The statistics of the CSQA dataset are summarized in Table \ref{dataset}.

Same as the pre-training stage, the input data for fine-tuning BERT\_CS is formed by concatenating each question-answer pair as a sequence. The hidden representations over the [CLS] token are run through a softmax layer to create the predictions.
The objective function is the same as Equations \ref{objective_function} and \ref{prob}.
We fine-tune BERT\_CS on CSQA for 2 epochs with a learning rate of 1e-5 and a batch size of 16. Table \ref{tab:CSQA_results} shows the accuracy on the CSQA test set from the baseline BERT models, the previous SOTA model CoS-E \citep{rajani2019explain}, and our BERT\_CS models. 
CoS-E model requires a large amount of human efforts to collect the Common Sense Explanations (CoS-E) dataset. 
In comparison, our multi-choice QA dataset $\mathcal{D}_{AMS}$ is constructed automatically. The BERT\_CS models significantly outperform the baseline BERT models with BERT\_CS$_{large}$ achieving 5.5\% absolute gain over the baseline BERT$_{large}$ model and 4.0\% absolute gain over the previous SOTA CoS-E model.

\begin{table*}[t!]
	\begin{center}
		\small
		\begin{tabular}{p{2.2cm}|c|c|c|c|c|c|c|c}
			\toprule
            \textbf{Model}             & \textbf{MNLI-(m/mm)} & \textbf{QQP} & \textbf{QNLI} & \textbf{SST-2} & \textbf{CoLA} & \textbf{STS-B} & \textbf{MRPC} & \textbf{RTE} \\
            \midrule
            BERT$_{base}$     & 84.6/83.4   & 71.2 & 90.5 & 93.5& 52.1 & 85.8 & 88.9 & 66.4  \\
            \midrule
            BERT\_triple$_{base}$     & 83.8/82.6   & 70.5 & 89.9 & 92.9& 49.6 & 85.3 & 88.7 & 65.1  \\
            \midrule
            BERT\_CS$_{base}$  & 84.7/83.9  &  72.1  &91.2 & 93.6& 54.3 & 86.4 & 85.9 & 69.5  \\
            \midrule
            BERT$_{large}$    & \textbf{86.7}/\textbf{85.9} &\textbf{72.1}& \textbf{92.7}& \textbf{94.9} & 60.5 & 86.5 & \textbf{89.3} & 70.1 \\
            \midrule
            BERT\_CS$_{large}$    & \textbf{86.7}/85.8 &\textbf{72.1}& 92.6& 94.1 & \textbf{60.7} & \textbf{86.6} & 89.0 & \textbf{70.7} \\
			\bottomrule
		\end{tabular}
	\end{center}
	\caption{The results of different models on the GLUE test sets. We use the same measure criterion as \citet{devlin-etal-2019-bert}. BERT\_CS$_{large}$ achieves comparable performance with BERT$_{large}$ and BERT\_CS$_{base}$ slightly better performance than BERT$_{base}$, verifying that our multi-choice QA based pre-training approach can maintain the performance on common NLP tasks.}
	%We report Matthews corr. on CoLA, Spearman corr. on STS-B, accuracy on MNLI, QNLI, SST-2 and RTE, F1-score on QQP and MRPC, which is the same as \cite{devlin-etal-2019-bert}.}
	\label{tab:GLUE_results}
\end{table*}

\subsection{Winograd Schema Challenge}
The WSC task \citep{levesque2012winograd} is introduced for testing AI agents for commonsense knowledge and is considered one of the most difficult commonsense reasoning datasets \citep{zhou2019evaluating}.
WSC consists of 273 instances for pronoun disambiguation. 
For an example sentence ``The delivery truck zoomed by the school bus because \textbf{it} was going so fast.'' and a corresponding question ``What does the word \textbf{it} refers to?'', the machine is expected to answer ``delivery truck'' instead of ``school bus''. 

We follow \citet{kocijan2019surprisingly} and employ the WSCR dataset \citep{rahman-ng-2012-resolving} as the training data. The WSCR dataset is partitioned into a training set of 1,322 examples and a test set of 564 examples. We use the WSCR training partition for fine-tuning pre-trained BERT\_CS models (Section~\ref{sect:pre-training}) and the WSCR test partition for validating BERT\_CS models, respectively, and test the fine-tuned BERT\_CS models on the WSC dataset.  We transform the pronoun disambiguation problem into a multi-choice QA problem. We mask the pronoun word with a special token [QW] to construct a question, and put the two candidate phrases as candidate answers.
The remaining procedures are the same as the CSQA task.
We use the same loss function as \citet{kocijan2019surprisingly}. 
That is, if c$_1$ is correct and c$_2$ is not, the loss is
\begin{equation}
\begin{aligned}
L = &- {\rm log}p(c_1|s) + \\
    &\alpha \cdot max(0, {\rm log}p(c_2|s)-{\rm log}p(c_1|s)+\beta), %\notag
\end{aligned}
\end{equation}
where $p(c_1|s)$ follows Equation \ref{prob} with $N=2$, $\alpha$ and $\beta$ are hyper-parameters.
Similar to \citet{kocijan2019surprisingly}, we search $\alpha \in \{2.5,5,10,20\}$ and $\beta \in \{0.05,0.1,0.2,0.4\}$ by optimizing the accuracy on the WSCR test partition (i.e., the development set for the WSC dataset).
We set the batch size 16 and the learning rate 1e-5.
We evaluate our models on the WSC dataset and its various partitions defined in \citet{trichelair-etal-2019-reasonable}. 
We also evaluate the fine-tuned BERT\_CS model (without using the WNLI training data for further fine-tuning) on the WNLI test set, one of the GLUE tasks \citep{wang-etal-2018-glue}.
We first transform the examples in WNLI from the premise-hypothesis format into the pronoun disambiguation problem format \citep{kocijan2019surprisingly} and then transform it into the multi-choice QA format.

The results on the WSC dataset and its various partitions and the WNLI test set are shown in Table~\ref{tab:WSC_results}. Note that \emph{non-assoc.} denotes instances in WSC that no antecedent is statistically preferred, and \emph{assoc.} denotes the rest instances. Columns \emph{unswitched} and \emph{switched} denote the accuracy on the unswitched and switched switchable subset defined in \citep{trichelair-etal-2019-reasonable}, and \emph{consist.} denotes the percentage of predictions that change after candidates in the switchable subset are switched. 
Higher scores for \emph{non-assoc.} and \emph{const.} suggest a better commonsense reasoning capability \citep{trichelair-etal-2019-reasonable}.
Note that the results for \citet{ruan2019exploring} are fine-tuned on the whole WSCR dataset, including its training and test set partitions.
Results for Ensemble 14 LMs \citep{trinh2018simple} and Knowledge Hunter\citep{emami2018knowledge} are cited from \citet{trichelair-etal-2019-reasonable}.
Results for ``BERT$_{large}$ + MTP" is cited from \citet{kocijan2019surprisingly} as the baseline of applying BERT to the WSC task. 
As can be seen from the table, our ``BERT\_CS$_{large}$ + MCQA" achieves the best performance on all of the evaluation criteria except being the second best on \emph{consist.}\footnote{Knowledge Hunter achieves a better \emph{consist.} score by a rule-based method \citep{trichelair-etal-2019-reasonable}.}, and achieves a 3.3\% absolute improvement on the WSC dataset over the previous SOTA results from \citet{kocijan2019surprisingly}. Further analysis shows that ``BERT$_{large}$ + MCQA" achieves better performance than ``BERT$_{large}$ + MTP" on the WSC and WNLI testsets and achieves significant improvements on \emph{consist.}, suggesting that MCQA may be a better problem formatting method than MTP for the WSC task. Since the CSQA dataset is created using concepts from ConceptNet and $\mathcal{D}_{AMS}$ is created using ConceptNet, $\mathcal{D}_{AMS}$ may be considered as an augmented data for CSQA\footnote{We will compare the BERT\_CS results in Table~\ref{tab:CSQA_results} with fine-tuning the original BERT on $\textrm{CSQA-train}+\mathcal{D}_{AMS}$.}. However, these WSC results demonstrate that the proposed approach is general for improving commonsense reasoning capabilities of the pre-trained models.

\begin{table*}[t!]
	\begin{center}
		\small
		\begin{tabular}{c|p{2.5cm}|c|c|c}
			\toprule
			\textbf{No.} &
            \textbf{Model}      & \textbf{Source Data}  & \textbf{Tasks}       & \textbf{Accuracy}  \\
            \midrule
            1 & BERT   & -  & -    & 58.2     \\
            \midrule
            2 & BERT\_triple & ConceptNet & MCQA    & 59.1     \\
            \midrule
            3 & BERT\_CS\_random  & Wikipedia and ConceptNet  &  MCQA     & 59.4 \\
            \midrule
            4 & BERT\_CS\_MLM    & Wikipedia and ConceptNet & MCQA+MLM              & 59.9  \\
            \midrule
            5 & BERT\_MLM    & Wikipedia and ConceptNet & MLM              & 58.8  \\
            \midrule
            6 & BERT\_CS & Wikipedia and ConceptNet & MCQA  & \textbf{60.8}  \\
			\bottomrule
		\end{tabular}
	\end{center}
	\caption{Ablation analysis: Model accuracy (\%) from different pre-training strategies on the CSQA development set. The source data and pre-training tasks are employed to pre-train BERT\_CS. \textbf{MCQA} denotes the multi-choice question answering task and \textbf{MLM} denotes the masked language modeling task.}
	\label{tab:analysis}
\end{table*}

\subsection{GLUE}
The GLUE benchmark \citep{wang-etal-2018-glue} is a collection of diverse natural language understanding tasks, including single-sentence tasks CoLA and SST-2, similarity and paraphrasing tasks MRPC, STS-B and QQP, and natural language inference tasks MNLI, QNLI, RTE and WNLI.
% The General Language Understanding Evaluation (GLUE) benchmark \cite{wang-etal-2018-glue} is a collection of diverse natural language understanding tasks, including MNLI \cite{williams-etal-2018-broad}, QQP \cite{QQP}, QNLI \cite{rajpurkar-etal-2016-squad}, SST-2 \cite{socher-etal-2013-recursive}, CoLA \cite{warstadt2018neural}, STS-B \cite{cer-etal-2017-semeval}, MRPC \cite{dolan-brockett-2005-automatically}, RTE \cite{bentivogli2009fifth} and WNLI \cite{levesque2012winograd}, of which CoLA and SST-2 are single-sentence tasks, MRPC, STS-B and QQP are similarity and paraphrase tasks and MNLI, QNLI, RTE and WNLI are natural language inference tasks.
To investigate whether our pre-training approach can maintain the performance on common text classification and NLI tasks, we evaluate BERT\_CS on 8 GLUE datasets and compare the performance with the baseline BERT models.
Following \citet{devlin-etal-2019-bert}, we use batch size 32 and fine-tune for 3 epochs for all GLUE tasks, and select the fine-tuning learning rate among 1e-5, 2e-5, and 3e-5 based on the performance on the development set.
Results are presented in Table~\ref{tab:GLUE_results}. 
We observe that BERT\_CS$_{large}$ achieves comparable performance with BERT$_{large}$ and BERT\_CS$_{base}$ achieves slightly better performance than BERT$_{base}$.  We hypothesize that the commonsense knowledge may not be required for the GLUE tasks. On the other hand, these results demonstrate that our proposed pre-training approach does not degrade the language representation capabilities of BERT models. 

\section{Analysis}
\subsection{Pre-training Strategies}
We conduct several experiments investigating different data creation approaches and pre-training tasks on the BERT$_{base}$ model.
For simplicity, we discard the subscript $base$ in this subsection.

In order to compare the efficacy of our data creation approach versus the data creation approach in \citet{sun2019probing},
same as \citet{sun2019probing}, we collect 606,564 triples from ConceptNet and construct 1,213,128 questions, each with a correct answer and four distractors. This dataset is denoted the TRIPLES dataset.
We pre-train BERT models on the TRIPLES dataset with the same hyper-parameters as the BERT\_CS models and the resulting model is denoted BERT\_triple.

We create several other model counterparts:
First, distractors are formed by randomly picking concept$_1$ or concept$_2$ in ConceptNet instead of those sharing the same concept$_2$ or concept$_1$ and the same relation with the correct answers. We denote the resulting model from this dataset BERT\_CS\_random.
Second, we randomly mask 15\% WordPiece tokens \citep{wu2016google} of the question as in \citet{devlin-etal-2019-bert} and then conduct both multi-choice QA task and MLM task simultaneously.  The resulting model is denoted BERT\_CS\_MLM.
Third, instead of pre-training BERT with a multi-choice QA task that chooses the correct answer from several candidate answers, we mask concept$_1$ and concept$_2$ and pre-train BERT with the MLM task. We denote the resulting model from this pre-training task BERT\_MLM.

% \begin{itemize}
% \item Distractors are formed by randomly picking concept$_1$ or concept$_2$ in ConceptNet instead of those sharing the same concept$_2$ or concept$_1$ and the same relation with the correct answers. We denote the resulting model from this dataset BERT\_CS\_random.
% \item Instead of pre-training BERT with a multi-choice QA task that chooses the correct answer from several candidate answers, we mask concept$_1$ and concept$_2$ and pre-train BERT with a masked language model (MLM) task. We denote the resulting model from this pre-training task BERT\_MLM.
% \item We randomly mask 15\% WordPiece tokens \cite{wu2016google} of the question as in \cite{devlin-etal-2019-bert} and then conduct both multi-choice QA task and MLM task simultaneously.  The resulting model is denoted BERT\_CS\_MLM.
% \end{itemize}

All these BERT models are fine-tuned on the CSQA training set with the same hyper-parameters as described in Section~\ref{subsect:csqa} and the results are shown in Table~\ref{tab:analysis}.
%We observe the following from Table~\ref{tab:analysis}.
Comparing model 1 and model 2, we find that pre-training on ConceptNet benefits the CSQA task even with the triples as input instead of sentences. 
Further comparing model 2 and model 6, we find that constructing natural language sentences as input for pre-training BERT performs better on the CSQA task than pre-training using triples. 
We also conduct more detailed comparisons between fine-tuning model 1 and model 2 on GLUE tasks. 
The results are shown in Table \ref{tab:GLUE_results}.
BERT\_triple$_{base}$ yields much worse results than BERT$_{base}$ and BERT\_CS$_{base}$, demonstrating that pre-training directly on triples may hurt the sentence representation capabilities of BERT. 

Comparing model 3 and model 6, we find that pre-training BERT benefits from a more difficult dataset.  In our selection method, all candidate answers share the same (concept$_1$, relation) or (relation, concept$_2$), that is, these candidates have close meanings. These more confusing candidates force BERT\_CS to distinguish meanings of synonyms, resulting in a more powerful BERT\_CS model.
Comparing model 5 and model 6, we find that the multi-choice QA task works better than the MLM task as the pre-training task for the target multi-choice QA task. 
We hypothesize that, for the MLM task, BERT is required to predict each masked wordpiece (in concepts) independently; whereas, for the multi-choice QA task, BERT is required to model the whole concepts instead of paying much attention to a single wordpiece. Particularly, our approach creates semantically confusing distractors and benefits incorporating commonsense knowledge and learning deep semantics.
Comparing model 4 and model 6, we observe that adding the MLM task hurts the performance of BERT\_CS. This is probably because masking 15\% words in questions cause the questions to be far from natural language sentences hence cause a pretrain-finetune discrepancy \citep{yang2019xlnet}. In contrast, the MCQA task only masks one concept, which relieves this discrepancy. We hypothesize with these advantages of MCQA over MLM, BERT\_CS does not degrade the language representation capabilities of BERT models. 
Among all the models in this analysis, the proposed BERT\_CS achieves the best performance on the CSQA development set.

\begin{table*}[t!]
	\begin{center}
		\small
		\begin{tabular}{p{8cm}|p{1.95cm}|c|c}
			\toprule
            \textbf{Question}   & \textbf{Candidates}  & \textbf{BERT}$_{large}$       & \textbf{BERT\_CS}$_{large}$  \\
            \midrule
            1) Dan had to stop Bill from toying with the injured bird. [\textbf{He}] is very compassionate.
            &A) \textbf{Dan}  B) Bill 
            & B & \textbf{A} \\
            \midrule
            2) Dan had to stop Bill from toying with the injured bird. [\textbf{He}] is very cruel.
            &A) Dan B) \textbf{Bill}
            & \textbf{B} & \textbf{B} \\
            \midrule
            3) The trophy doesn't fit into the brown suitcase because [\textbf{it}] is too large.
            &A) \textbf{the trophy} B) the suitcase
            & B    & B  \\
            \midrule
            4) The trophy doesn't fit into the brown suitcase because [\textbf{it}] is too small.   
            &A) the trophy B) \textbf{the suitcase}
            & A    & A  \\
			\bottomrule
		\end{tabular}
	\end{center}
	\caption{Several cases from the WSC dataset. The pronouns in questions are in square brackets. The correct answers and correct model predictions are in boldface.}
	\label{tab:case_study}
\end{table*}

\subsection{Performance Curve}
\begin{figure}[!t]
	\centering
	\includegraphics[width=2.7in]{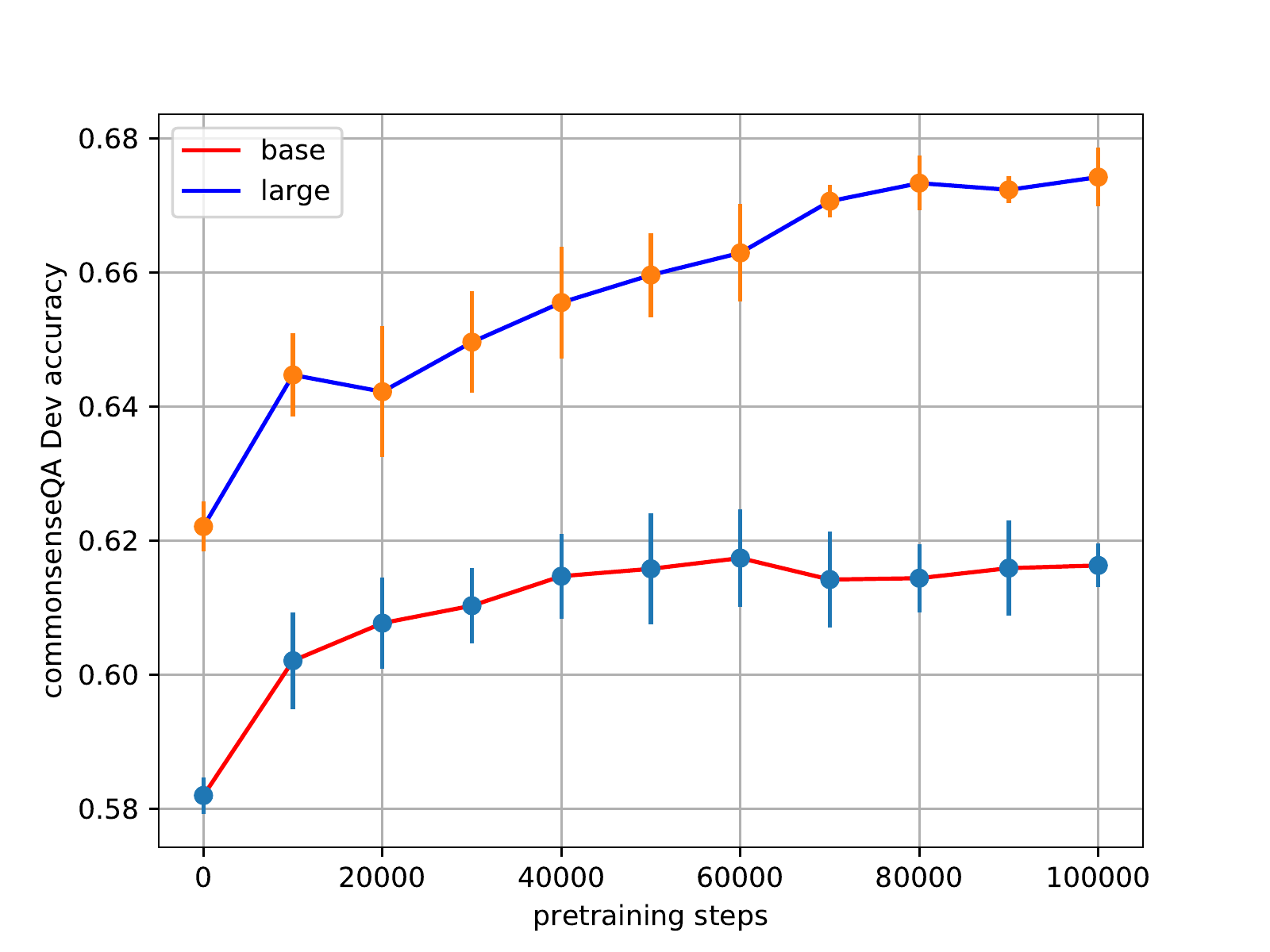}
	\caption{BERT\_CS$_{base}$ and BERT\_CS$_{large}$ accuracy on the CSQA development set against the number of pre-training steps.}
	\label{fig:performance_curve}
\end{figure}

We investigate the performance from BERT\_CS on the CSQA development set against the pre-training steps.
For every 10,000 training steps, we save the model as the initial model for fine-tuning. 
For each model, we run experiments for 10 times with the same pre-trained checkpoint but use different random seeds.
Due to instability of fine-tuning BERT~\citep{devlin-etal-2019-bert}, we remove the results that are significantly lower than the mean.
%\footnote{In our experiments, we remove the accuracy lower than 0.57 for BERT\_CS$_{base}$ and 0.60 for BERT\_CS$_{large}$.}.
Figure~\ref{fig:performance_curve} shows the mean and standard deviation of accuracy. The performance of BERT\_CS$_{base}$ converges around 50,000 training steps while BERT\_CS$_{large}$ still improves at 100,000 steps, suggesting that BERT\_CS$_{large}$ is probably more powerful for incorporating commonsense knowledge.
We also observe that pre-training with 2 epochs produces worse performance than with 1 epoch, probably due to over-fitting.
Pre-training with more QA samples may benefit the BERT\_CS models and we will explore this in the future work.

\subsection{Error Analysis}
Table \ref{tab:case_study} shows several cases from the WSC dataset. 
Questions 1 and 2 only differ in the words ``compassionate" and ``cruel".
BERT\_CS$_{large}$ chooses correct answers for both questions while BERT$_{large}$ chooses the same choice ``Bill" for both questions. 
We hypothesize that BERT$_{large}$ tends to choose the closer candidates. To investigate this hypothesis, we split the WSC test set into CLOSE and FAR subsets, based on whether the correct answer is closer or farther to the pronoun than another candidate.
We find that BERT\_CS$_{large}$ achieves the same 82.4\% accuracy on the CLOSE set as BERT$_{large}$ and significantly better accuracy on the FAR set than BERT$_{large}$ (68.6\% versus 60.6\%), suggesting that BERT\_CS$_{large}$ is probably less influenced by proximity and more focused on semantics. Questions 3 and 4 only differ in ``large" and ``small".
Neither BERT\_CS$_{large}$ nor BERT$_{large}$ chooses the correct answers, probably due to ``suitcase is large" and ``trophy is small" being frequent in the pre-training text. Next, we plan to explore reducing the sensitivity to language models.

\section{Conclusion}
We propose a pre-training approach for incorporating commonsense knowledge into language representation models and a method for automatically constructing a multi-choice QA dataset for pre-training.  Experiments demonstrate that the proposed approach significantly outperforms SOTA on commonsense-related CSQA and WSC tasks, while maintaining comparable performance on GLUE tasks to the BERT models. In future work, we will use it to incorporate commonsense knowledge into models such as XLNet \citep{yang2019xlnet} and RoBERTa \citep{liu2019roberta}.
%, and exploit other KGs such as Freebase \citep{bollacker2008freebase}.

\bibliography{acl2020,anthology}
\bibliographystyle{acl_natbib}

\end{document}